\documentclass[10pt,twocolumn,letterpaper]{article}

\usepackage[pagenumbers]{cvpr} % To force page numbers, e.g. for an arXiv version

\usepackage{lineno}

\usepackage[dvipsnames]{xcolor}

\definecolor{cvprblue}{rgb}{0.21,0.49,0.74}
\usepackage[pagebackref,breaklinks,colorlinks,citecolor=cvprblue]{hyperref}

\def\paperID{10065} % *** Enter the Paper ID here
\def\confName{CVPR}
\def\confYear{2024}

\title{Generating floorplans for various building functionalities via latent diffusion model}

\author{
  Mohamed R. Ibrahim\textsuperscript{*,1},
  Josef Musil\textsuperscript{2},
  Irene Gallou\textsuperscript{2}
  \\
  \textsuperscript{1}University of Leeds, Leeds, UK\\
  \textsuperscript{2}SMG, Foster + Partners, London, UK
  \\
  \textsuperscript{*}Corresponding author: \texttt{geomi@leeds.ac.uk}
}

\begin{document}
\maketitle

\begin{abstract}

In the domain of architectural design, the foundational essence of creativity and human intelligence lies in the mastery of solving floorplans, a skill demanding distinctive expertise and years of experience. Traditionally, the architectural design process of creating floorplans often requires substantial manual labour and architectural expertise. Even when relying on parametric design approaches, the process is limited based on the designer's ability to build a complex set of parameters to iteratively explore design alternatives. As a result, these approaches hinder creativity and limit discovery of an optimal solution. Here, we present a generative latent diffusion model that learns to generate floorplans for various building types based on building footprints and design briefs. The introduced model learns from the complexity of the inter-connections between diverse building types and the mutations of architectural designs. By harnessing the power of latent diffusion models, this research surpasses conventional limitations in the design process. The model's ability to learn from diverse building types means that it cannot only replicate existing designs but also produce entirely new configurations that fuse design elements in unexpected ways. This innovation introduces a new dimension of creativity into architectural design, allowing architects, urban planners and even individuals without specialised expertise to explore uncharted territories of form and function with speed and cost-effectiveness.

\end{abstract}

\begin{figure}[!ht]
  \centering

   \includegraphics[width=1\linewidth]{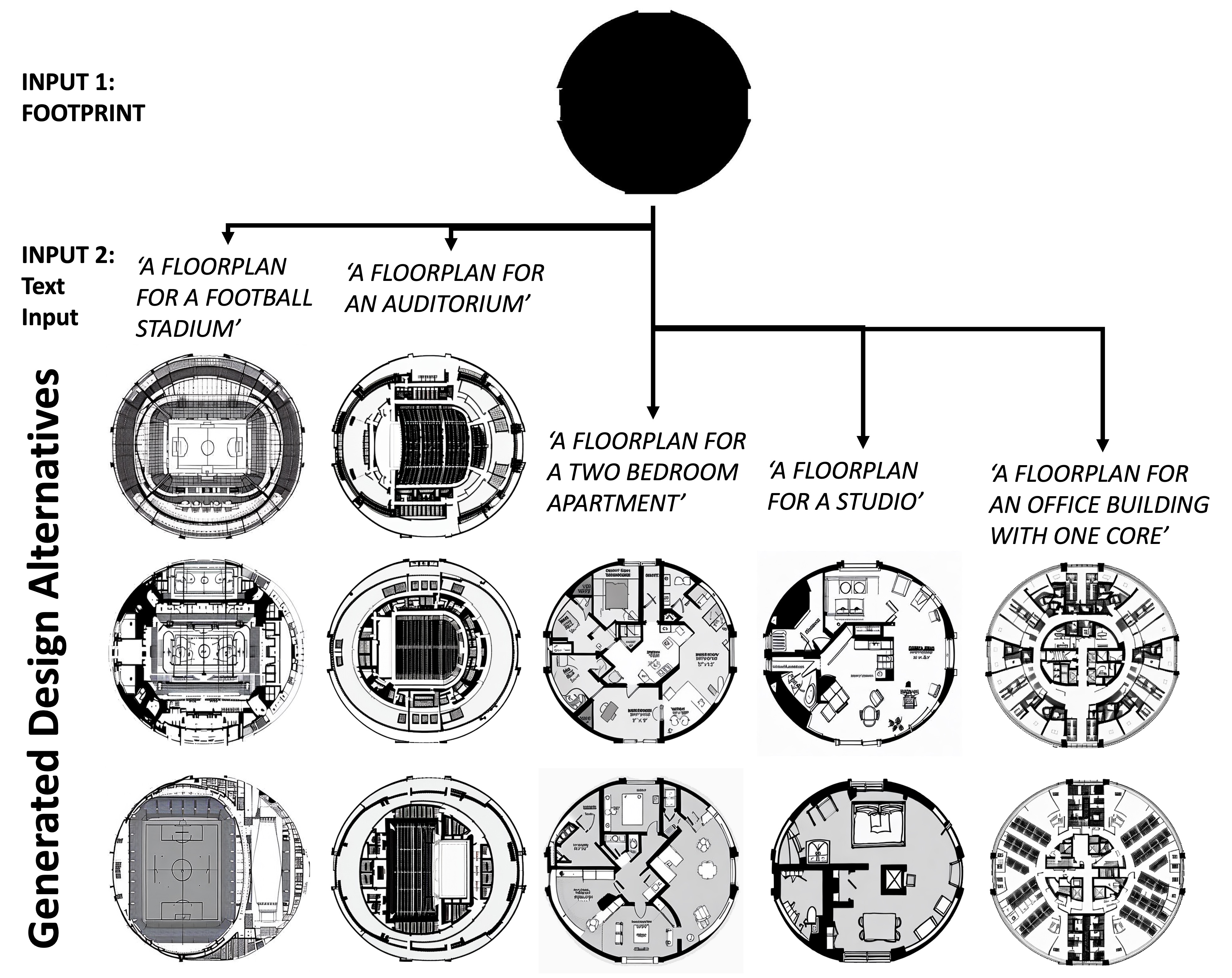}

   \caption{Generated floorplans illustrating various building types. Our model takes two inputs—lot footprint and textual description—and outputs the corresponding architectural floorplans.}
   \label{fig:fig1}
\end{figure}

\section{Introduction}

The process of architectural design is a complex and multifaceted endeavour that presents numerous challenges \cite{archprocess_1, archprocess_2, archprocess_3, archprocess_4}. Architects face the task of finding optimal solutions that not only satisfy the functional requirements of a building but also possess aesthetic appeal, adhere to site regulations, and meet the expectations of their clients. Ultimately, architectural design goes beyond mere functionality and compliance; it aspires to make a lasting impression on the built environment. Architects aspire to create designs that transcend time, leaving a legacy that enriches the fabric of a place. Achieving functionality in architectural design, however, involves a deep understanding of the intended purpose and user needs which require an intensive iterative design process. To reach an optimal design, accordingly, the architectural design process of creating floorplans often requires substantial manual labour and architectural expertise which are often limited by projects’ resources and timelines. Over the last few decades, there has been continuous progress in design process and aided software developments to explore iterative design alternatives before decision-making, moving from hand sketching and drawing to Computer Aided Drawing (CAD) software, to parametric design, a realm where designers develop set of rules and parameters, orchestrating a design process that is both guided and adaptive \cite{parametric_1,parametric_2, parametric_3, parametric_4, parametric_5}. Despite the utilisation of parametric design methodologies, the design process can encounter constraints rooted in the designer's ability to construct complex parameter sets for the iterative exploration of design alternatives. This dynamic invariably imposes limitations on creative ingenuity and restricts the potential for developing an optimal solution.

Within the remarkable strides witnessed in Artificial Intelligence (AI) and generative models \cite{goodfellow_generative_2014,diffusion_paper, latent_diffusion, gan_1, gan_10, gan_13}, the realm of design has undergone a paradigm shift, enabling the exploration of design possibilities with speed and minimal resource allocation. While an increasing number of tools now exists for the generation of images \cite{latent_diffusion, gan_12, gan_13, gan_14}, a notable gap remains in the context of a dedicated framework tailored specifically for architectural design and floorplans. To the best of our current knowledge, a single deep model that autonomously generates floorplans across diverse building types is yet to be realised. 

In response to this research gap, we present an innovative model poised to revolutionise the process of generating floorplans for a wide spectrum of building typologies. The key feature of the introduced model lies in its inherent scale-agnostic capability (See Fig. \ref{fig:fig1}). This newly introduced attribute empowers the model to fluidly adapt the scale of its generated floorplans in direct response to provided footprint and design brief inputs, thereby guaranteeing a harmonious alignment of the generated designs with the specific design needs of diverse building typologies. This paper makes three distinctive contribution as follows: 1) This research introduces a novel dataset tailored explicitly for generating floorplans. This dataset is uniquely structured to facilitate training models using two multi-modal inputs (text, and image), transcending the confines of conventional datasets, 2) the cornerstone of this research lies in the unveiling of a new generative model. This model is designed not only to empower architects and planners but also to enable individuals without specialized architectural expertise to explore novel concepts of form and function. The model's intrinsic capabilities are harnessed to expedite creative explorations, all while balancing efficiency and cost-effectiveness, and 3) the empirical underpinning of this research is further fortified by an exhaustive ablation study, conducted to unravel the intricate nuances of design mutations within floorplans. This comprehensive analysis illuminates the model's capacity to evolve and adapt across various iterations, thereby providing a nuanced understanding of the dynamic interplay between design elements, 3) developing tools and incorporating gamification strategies to scale up the evaluation of AI-generated images using specialised human expertise. 

\section{Background}

The stated issue of generating floor plans pertains to four distinct domain areas that hold significance. In these areas, we will succinctly outline the cutting-edge advancements as well as the existing limitations.

\textbf{Iterative design: }Iterative design is a design methodology that applies to various fields, including the classical approach of designing floor plans \cite{iterative_1, iterative_2, iterative_3, iterative_4, iterative_5, iterative_6,iterative_7}. It comprises a cyclical approach of prototyping, testing, and refining to develop optimal floor plan designs. The process recognises the importance of ongoing feedback and continuous improvement throughout the design journey. When designing floor plans, the iterative approach begins by establishing design goals and understanding the requirements of the space and its occupants. Based on this information, an initial floor plan prototype is created, considering factors such as functionality, spatial organization, and aesthetics. The prototype is then evaluated through various means, such as reviewing user feedback, conducting usability tests, and consulting with experts in architecture and interior design. This evaluation stage helps identify strengths, weaknesses, and areas for improvement in the floor plan design.  Although iterative parametric design is a successful process for creating floor layouts \cite{parametric_1,parametric_2,parametric_3}, it has several drawbacks. The lack of resources and time is one of the major restrictions. The floor plan must be developed, tested, and improved with each step of the design process. As a result, there could be practical restrictions on how many iterations can be successfully completed within a certain timeframe or using the resources at hand. Additionally, not every design alternative might be investigated during the iterative design process. Instead of exhaustively examining every design option, architects and designers may need to concentrate on a subset of feasible solutions due to time restrictions or other practical concerns. This limitation means that there is a possibility of missing out on alternative design approaches or potentially superior solutions.

\textbf{Conditional image generation:} To explore different design strategies and layout options, relying on generative AI approaches for image generations becomes a crucial approach. Generative AI refers to a branch of artificial intelligence that focuses on creating models and algorithms capable of generating new and original content, such as images, text, audio, or even videos \cite{gan_1,gan_10, goodfellow_generative_2014, gan_review}. Generative AI has been particularly successful in the field of image generation. Models like Generative Adversarial Networks (GANs) have demonstrated the ability to generate highly realistic and diverse images. Conditional GANs allow for targeted image synthesis by conditioning the generation process on specific attributes or conditions, such as class labels, semantics, or textual descriptions \cite{gan_7,condition_gan_1, condition_gan_2, condition_gan_3, condition_gan_5, gan_review}. Similarly, few models have developed relying on conditional GANS to generate floor layouts from a zoning diagram \cite{floorplan_14,floorplan_15}. 
Lately, the field of image generation has witnessed remarkable advancements with the emergence of diffusion models \cite{diffusion_paper, latent_diffusion,controlnet, glide}. These models have demonstrated exceptional capabilities in generating high-resolution images from text prompts. Diffusion models are based on the concept of diffusion processes, which describe how an initial image evolves over time. In each step of the diffusion process, the model applies a series of transformations to the image, gradually reducing the noise and refining the details. By iteratively applying these transformations, diffusion models can generate high-quality images. To condition diffusion models for image generation, similar to conditional GANs, additional information can be provided to guide the generation process. This information can be in the form of conditioning variables, such as class labels or attributes, which influence the transformations applied during each diffusion step. The advantage of diffusion models lies in their ability to generate high-resolution and diverse images. By controlling the diffusion process and conditioning variables, users can have more control over the generated images, allowing for targeted and controlled image synthesis.

\textbf{Generative floorplans: } While numerous generative models have been developed to create residential floor plans \cite{floorplan_1, floorplan_2, floorplan_3, floorplan_4, floorplan_5, floorplan_6, floorplan_7, floorplan_8, floorplan_9, floorplan_10, floorplan_11, floorplan_12, floorplan_13, floorplan_14, floorplan_15, floorplan_16, floorplan_17, floorplan_18, floorplan_19, floorplan_20, floorplan_21}, there exist considerable limitations in their scope and adaptability. The primary focus of these existing methods is on generating residential floor plans. Some are restricted to encoding walls \cite{floorplan_3}, while others are centred on the semantics of zoning layouts \cite{floorplan_14, floorplan_15} or rely on knowledge graphs representing zoning areas \cite{floorplan_2, floorplan_7, floorplan_10, floorplan_13, floorplan_17, floorplan_18, floorplan_19, floorplan_20, floorplan_21}. This leaves much to be desired in terms of versatility, particularly in adapting these models to produce layouts for different building types like offices, stadiums, libraries, or auditoriums. Moreover, the success of these models often depends heavily on human expertise, especially from architects and designers, to furnish optimal zoning layouts. This reliance on expert input inhibits the models from autonomously determining the most efficient zoning strategy, which limits their scalability and complicates their usability for non-experts. Another drawback is the computational efficiency of these models. The complexity of architectural design often leads to high computational costs, particularly when considering additional features or constraints, such as natural light, ventilation, or local building codes. This makes these models less viable for real-time or large-scale applications.

This research aims to address these challenges by introducing novel methodologies that expand the applicability of generative models to various building types and allow for more autonomous and computationally efficient design generation.

\section{Methodology}

\begin{figure}[ht]
  \centering

   \includegraphics[width=1\linewidth]{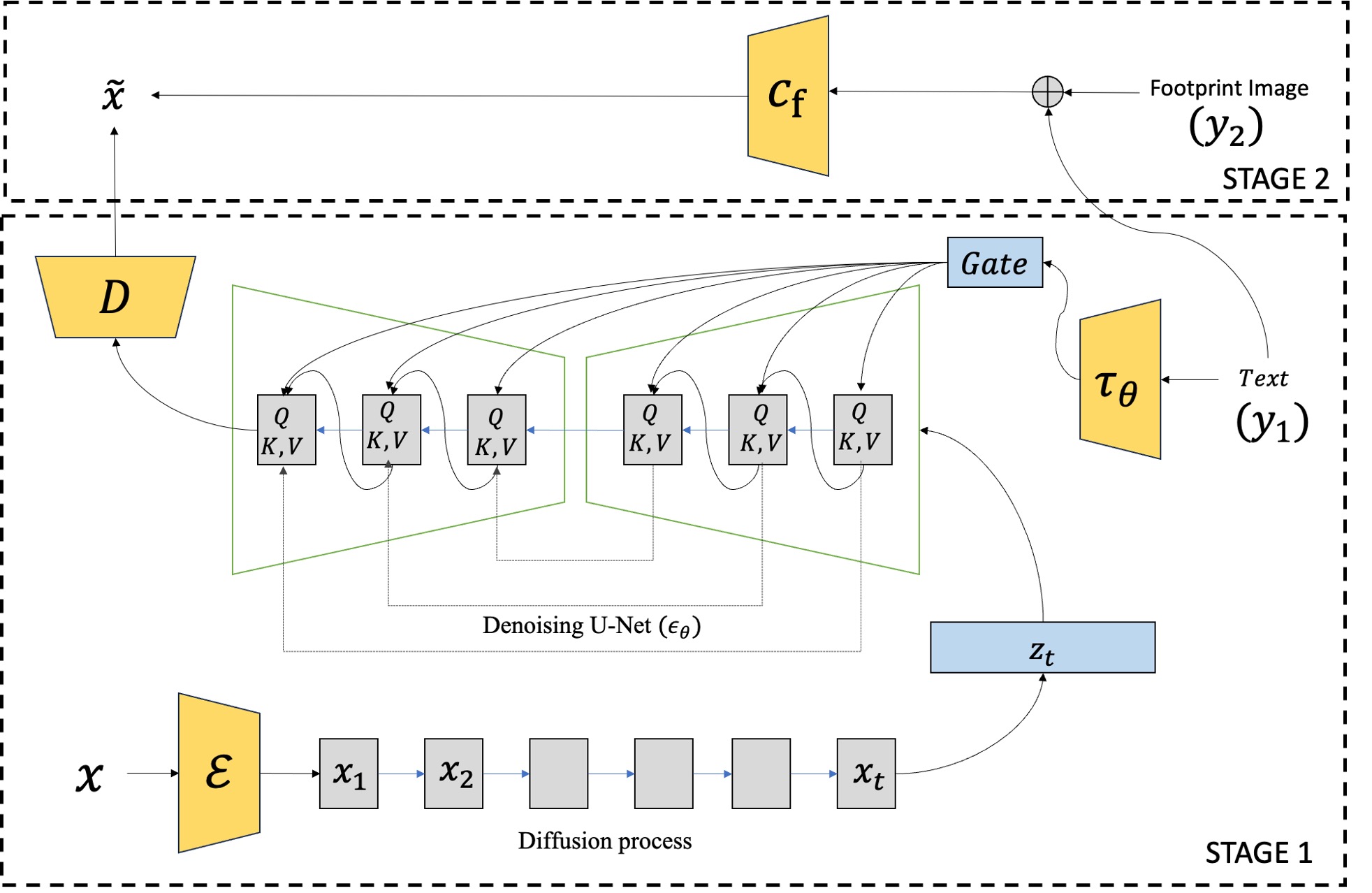}

   \caption{The overall proposed architecture based on the LDM model. }
   \label{fig:fig2}
\end{figure}

\subsection{Model Concept and Utility}

The concept of the proposed model is to allow planners and designers to generate a floorplan representation based on two given conditions: representing the building's footprint and a text briefing explaining the aimed design. The goal is to create a scale-agnostic model that, given the stated conditions, can adapt to generate a floorplan based on learning the specifics that correspond to a given building's type (i.e., residential, office building, stadium, etc.) without the scale being explicitly given. The main reason for creating a scale-agnostic model is to allow users to iteratively explore design options based on different architectural programs for a given footprint without worrying about calculating dimensionality and scale representation.

We propose a Latent Diffusion Model (LDM) that outputs an image \( \tilde{x} \in \mathbb{R}^{h \times w \times c} \) given two conditions: a design brief as a text prompt \( y_1 \) and a building's footprint in the form of an image \( y_2 \). The conditional distribution can be expressed in the form of \( p(x|y_1,y_2) \). To do so, the model architecture involves two main stages; a diffusion mode conditioned based on the first condition, and an additional training procedure to constrain and further condition the trained model in the first state based on the second condition. Fig.~\ref{fig:fig2} shows the overall architecture of the proposed model.

At the first stage, a given image sample \( x \) is passed through a diffusion process of \( t \) steps by introducing a Gaussian noise at each \( t \) step, given that \( t \) is sampled from \( \{1,\ldots,T\} \), to generate a latent space \( z_t \). $x$ is encoded to a latent space \( z_t \)  by a probabilistic encoder, and then it is decoded back to the original space by a probabilistic decoder. \( z_t \)  is represented as a probability distribution, typically Gaussian:

\begin{equation}
q_{\phi}(z_t|x) = \mathcal{N}(z; \mu(x), \sigma^2(x)I)
\end{equation}

where \( \mu(x) \) and \( \sigma^2(x) \) are the mean and variance of the Gaussian distribution and are functions of the input \( x \). They are learned by the encoder, \( I \) is the identity matrix and \( \phi \) represents the parameters of the encoder.

Afterwards, the latent space \( z_t \) alongside the text prompt representing the design brief is tokenized and fed forward to attention layers within an autoencoder to denoise the given sample.

Similar to \cite{controlnet}, we introduced an attention layer in the U-Net architecture, which is defined as:

\begin{equation}
{Attention} (Q,K,V) = {softmax} \left( \frac{QK^T}{\sqrt{d}} \right) \cdot V
\end{equation}

with:
\begin{equation}
Q=W_Q^{(i)} \cdot \phi_i(z_t), \quad K=W_K^{(i)} \cdot \tau_\theta(y), \quad V=W_V^{(i)} \cdot \tau_\theta(y)
\end{equation}

Given that \( \phi_i(z_t) \in \mathbb{R}^{N \times d_{\epsilon}^i} \) denotes a (flattened) intermediate representation of the U-Net implementing \( \epsilon_\theta \) and \( W_V^{(i)} \in \mathbb{R}^{d \times d_{\epsilon}^i} \), \( W_Q^{(i)} \in \mathbb{R}^{d \times d_{\tau}} \), \( W_K^{(i)} \in \mathbb{R}^{d \times d_{\tau}} \).

At the second stage, after training the first model, denoted as \( F(\cdot; \Theta) \) where \( \Theta \) represents a set of parameters that map an input \( x \) into \( \tilde{x} \), we freeze and clone \( \Theta \) in a trainable copy \( \Theta_c \), as introduced by \cite{controlnet} to avoid overfitting when a dataset is small while keeping the high fidelity of the large models trained at stage one. We connected the cloned network to a zero-convolution layer operation \( (1 \times 1) \) \( Z(\cdot; \cdot) \) where \( \tilde{x} \) is updated as an output of this network, given parameters \( \{\Theta_{z1}, \Theta_{z2}\} \) and the second condition \( y_2 \). The final output of the network can be expressed as:

\begin{equation}
\tilde{x} = F(x;\Theta) + Z(F(x+Z(y_2;\Theta_{z1});\Theta_c);\Theta_{z2})
\end{equation}

\subsection{Objective Loss}

In the first stage, the model focuses on learning both the temporal function $\tau_{\theta}$ and the noise model $\epsilon_{\theta}$ conditioned on a single context, such as text prompts \(y_1\). The loss function \(L_{LDM}\) that is optimized during this stage is given by:
\begin{equation}
L_{LDM} := \mathbb{E}_{E(x),y_1,\epsilon \sim \mathcal{N}(0,1),t} [\| \epsilon - \epsilon_\theta (z_t,t,\tau_\theta(y_1)) \|_2^2]
\end{equation}
where \(E(x)\) stands for the encoder function that maps the observed data \(x\) to the latent space, \(z_t\) is the latent variable at time \(t\), and \(\epsilon\) is sampled from a Gaussian distribution \(\mathcal{N}(0, 1)\).

In the second stage, the model learns to condition its noise model \(\epsilon_{\theta}\) on multiple contexts, namely the text prompts \(y_1\) and additional information like the building's footprint \(y_2\). The overall learning objective \(L\) for this stage is:
\begin{equation}
L = \mathbb{E}_{z_0,t,y_1,y_2,\epsilon \sim \mathcal{N}(0,1)} [\| \epsilon - \epsilon_\theta (z_t,t,y_1,y_2) \|_2^2]
\end{equation}

The difference between \(L\) and \(L_{LDM}\) is the inclusion of the second condition \(y_2\). Here, \(z_0\) denotes the initial latent state, \(z_t\) the latent state at time \(t\), and again \(\epsilon\) is sampled from \(\mathcal{N}(0, 1)\). This two-stage process allows the model to initially adapt to simpler conditions and then extend its capabilities to more complex, multiple conditional inputs. The model's parameters are jointly optimised to minimise the respective loss functions during each training stage. By using this approach, the conditional latent diffusion model aims to provide a nuanced representation that can be adapted based on various conditions, thereby increasing its versatility and applicability across different tasks and data types.

\section{Experiments and ablation study}
We conducted multiple experiments focused on enhancing image fidelity. The findings from these investigations are encapsulated in the models we present here, which exhibit superior performance in delivering high-quality images.

\begin{figure}[ht]
  \centering

   \includegraphics[width=1\linewidth]{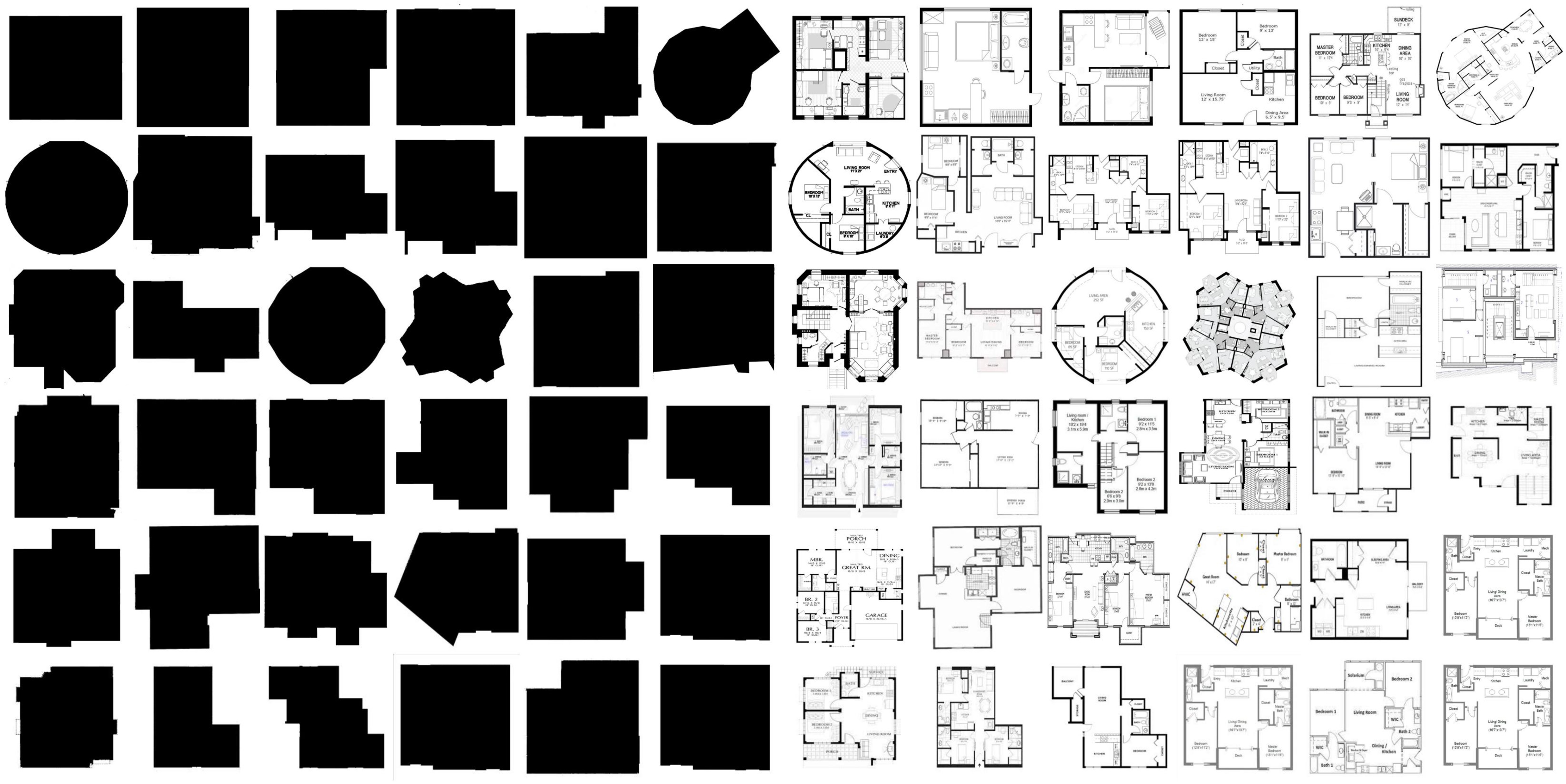}

   \caption{Examples of the introduced datasets, showing paired images and their corresponding text prompts.}
   \label{fig:data_fig}
\end{figure}

\textbf{Datasets:}
The first stage is trained on 400 million text-to-image synthesis using  LAION dataset \cite{laion400m}, whereas the second stage is trained in our newly collected dataset. While there are datasets that focuses on generating residdential floorplans from semantics \cite{floorplan_14}, to the best of our knowledge, currently, there is no existing dataset that encompasses all three elements of a textual description of design, footprint, and the final design of a floor plan. As a result, we took the initiative to create our own dataset, which consists of 500 pairs of images and corresponding text prompts. Each sample comprises three distinct components: 1) a mask image that visually represents the designated shape for the building's footprint, 2) a text prompt that serves as a concise brief for the intended floor plan design, and 3) a high-quality floor plan that represents the ultimate output of our model.
The first two components serve as the input for our model, while the third component represents the output generated by the model. To compile this dataset, we gathered floor plans from various sources on the internet. Additionally, we manually labelled each floor plan to accurately depict the corresponding footprint using mask images and to provide a concise description of the design based on the architectural expertise of our labellers. Fig. \ref{fig:data_fig} shows a diverse selection of data samples obtained from our dataset. It showcases the wide range of footprints alongside various building types, highlighting the variability within our collected data.

\textbf{Evaluation Metrics:}
We employed both quantitative and qualitative methods to assess the logic and excellence of the produced images \cite{gan_Evaluation, gan_review, latent_diffusion}. Although it's typical to calculate quantitative metrics like Fréchet Inception Distance (FID) \cite{fid}, Kernel Inception Distance (KID) \cite{kid, gan_Evaluation},  Structural Similarity Index (SSIM) \cite{SSIM} and Signal-to-Noise Ratio (PSNR) score \cite{PSNR}, the objective of producing multiple samples for identical inputs makes it challenging to evaluate the model quantitatively. In response to the challenge of lacking measurable criteria for evaluating the spatial layouts of the AI-generated floorplans, we devised an interactive, game-like method. This method assesses the architectural designs and qualities of these generated images through the judgment of architects who participate as players. We developed a straightforward tool that displays images, encompassing both authentic and generated examples. The participating architects (players) are tasked with rating each image on a scale from 1 to 10, with 10 being the highest, focusing exclusively on the composition and architectural integrity. We present a random selection of 30 images showcasing various types of buildings, with an equal mix of real and generated images. A crucial element of this game is its anonymity aspect; players are not informed which images are AI-generated and which are real, compelling them to base their evaluations solely on design composition. The link to the game: \href{https://floorplan.streamlit.app}{FloorplanGame} 

We also compare the results of our model to tools that are commonly utilised by architects and designers to aid in the design process of architectural projects. These tools include Stable Diffusion \cite{latent_diffusion,diffusion_paper} or DALLE-2 \cite{dalle2}. In this research, we will assess how our introduced model surpasses these tools in terms of its ability to understand technical terms within text prompts and the quality of the floor plans it generates.

\textbf{Model Performance: }
After training our model (see implementation details in Supplementary), Fig. \ref{fig:loss} shows the training and validation losses of the trained model over 549 training cycles (epochs), illustrating consistent performance throughout both training and validation phases. In Table \ref{tab:metrics}, we compare various image generative methods using multiple evaluation metrics. Notably, our model stands out as it achieves superior performance across all metrics. With the lowest FID of 22.436, our model exhibits remarkable similarity to real images. Additionally, it boasts the lowest KID of 1.844, indicating the smallest discrepancy between the generated and real image distributions. Our model also excels in structural similarity, as reflected in the highest SSIM score of 0.130.  It attains PSNR score of 7.596, demonstrating excellent image quality. 

\begin{figure}[ht]
  \centering

   \includegraphics[width=1\linewidth]{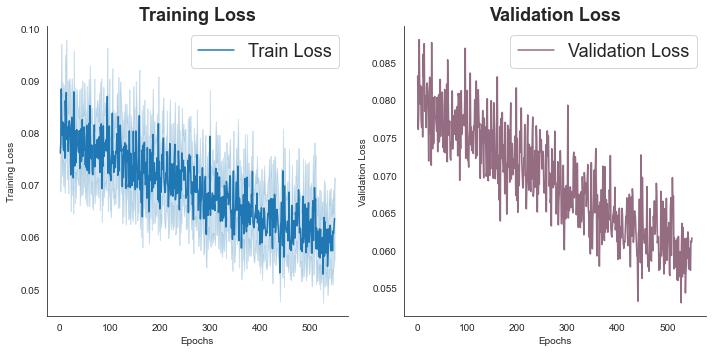}

   \caption{Training and validation losses for the introduced model.}
   \label{fig:loss}
\end{figure}

In Table \ref{tab:game-evaluation}, we present the outcomes of a game in which architects evaluated both real and generated floorplans. The t-test performed on these scores yielded statistical significance (p=0.001), indicating  a slightly notable differences between the two groups. Architects awarded an average score of 6.89 (±2.71 std) to real floorplans, with a minimum score of 1, a maximum score of 10, and a median score of 8. In contrast, generated floorplans received an average score of 5.36 (±2.91 std), with a minimum score of 0, a maximum score of 10, and a median score of 5.   Finally, in Figure \ref{fig:game}, we present the outcomes of human evaluations comparing the generated images against the real ones, without revealing their origins. The results illustrate the closeness of the scores between the generated and real images, with a slightly better score observed for the real images.

\begin{table}[ht]
\centering
\caption{Evaluation Metrics for Different image generative models}
\label{tab:metrics}
\begin{tabular}{@{}lccccc@{}}
\toprule
Method & FID $\downarrow$ & KID $\downarrow$ & SSIM $\uparrow$ & PSNR $\uparrow$\\ 
\midrule
LDM \cite{latent_diffusion} & 44.594 &  2.430 & 0.042 & 6.915 \\
DaLLE-2 \cite{dalle2} & 42.265 & 3.087 & 0.037 & 6.337  \\
Ours  & \textbf{22.436}  & \textbf{1.844} & \textbf{0.130} &  \textbf{7.596} \\
\bottomrule
\end{tabular}
\end{table}

\begin{table}[ht]
\centering
\caption{ The scores of all players (architects) of the Floorplan Evaluation Game (N=10). The t-test is statistically significant (p=0.001)}
\label{tab:game-evaluation}
\begin{tabular}{@{}lccccc@{}}
\toprule
Group & mean ± std & Min & Max & Median & t-test \\ 
\midrule
Real      & 6.89 ± 2.71  & 1 & 10 & 8 & - \\
Generated & 5.36 ± 2.91  & 0 & 10 & 5 & 3.917 \\
\bottomrule
\end{tabular}

\end{table}

\begin{figure}[!ht]
  \centering

   \includegraphics[width=1\linewidth]{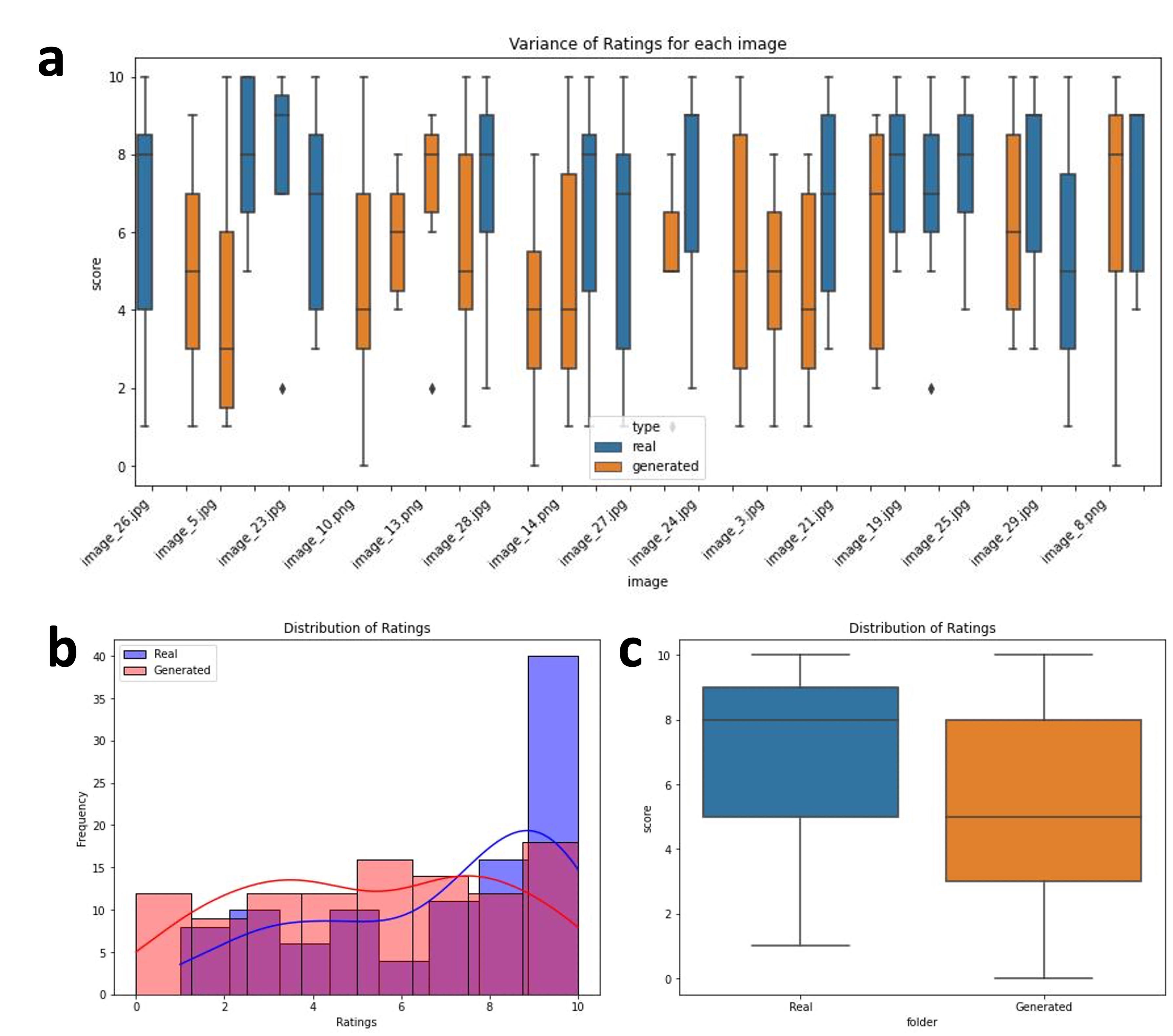}

   \caption{Results of human evaluations for real and generated images in the Floorplan Game. a) Boxplot displaying rating variations for each image. b) Distribution of scores (0-10) for both image types, showing score overlap. c) Boxplot comparing average scores across all players for real and generated images.}
   \label{fig:game}
\end{figure}

\begin{figure}[ht]
  \centering

   \includegraphics[width=1\linewidth]{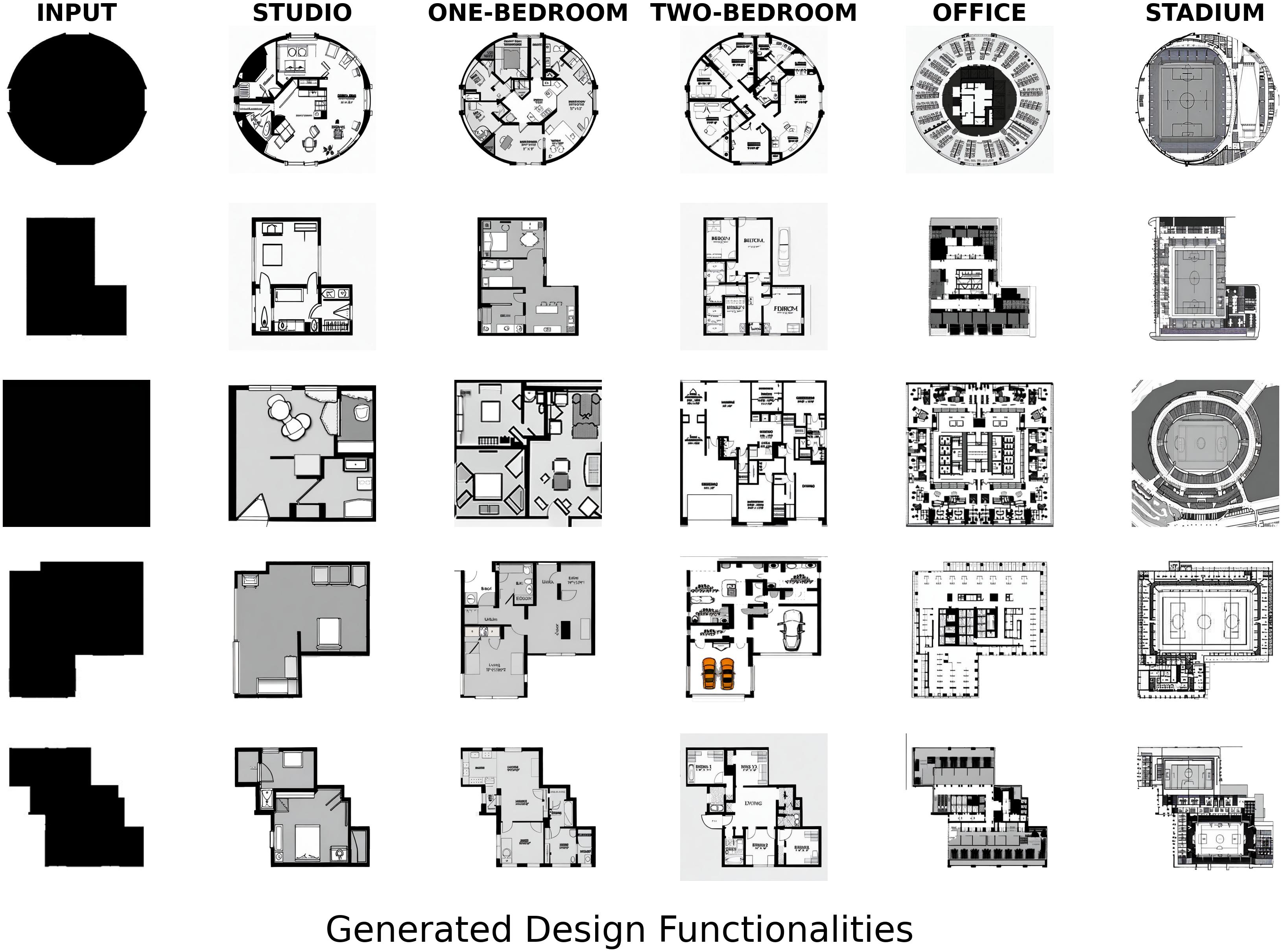}

   \caption{Generated floorplans from various input shapes for different design functionality}
   \label{fig:fig3}
\end{figure}

\textbf{Generating floorplans based on functionality: }
Investigating the outcomes of our trained model in Fig. \ref{fig:fig3}, we unveil the achievements in generating diverse floor plans catering to distinct design briefs and footprints. The visual representations presented therein not only underscore the remarkable versatility of our models but also highlight their innate capacity to seamlessly adapt and address the challenge of producing floorplans of identical footprints across a broad spectrum of design requirements. The demonstrated capacity to reconcile intricate design briefs with a consistent footprint engenders a new dimension of architectural flexibility, offering the potential to revolutionise how spaces are conceptualised and optimised with minimal resources.  Fig. \ref{fig:stadium},  \ref{fig:office}, \ref{fig:apartment}, \ref{fig:library}, and \ref{fig:auditorium} show further generated designs for various buildings footprints for stadium, office, apartment, library and auditorium buildings respectively highlighting the complexity and diversity of the generated design alterantives.

\begin{figure}[ht]
  \centering

   \includegraphics[width=1\linewidth]{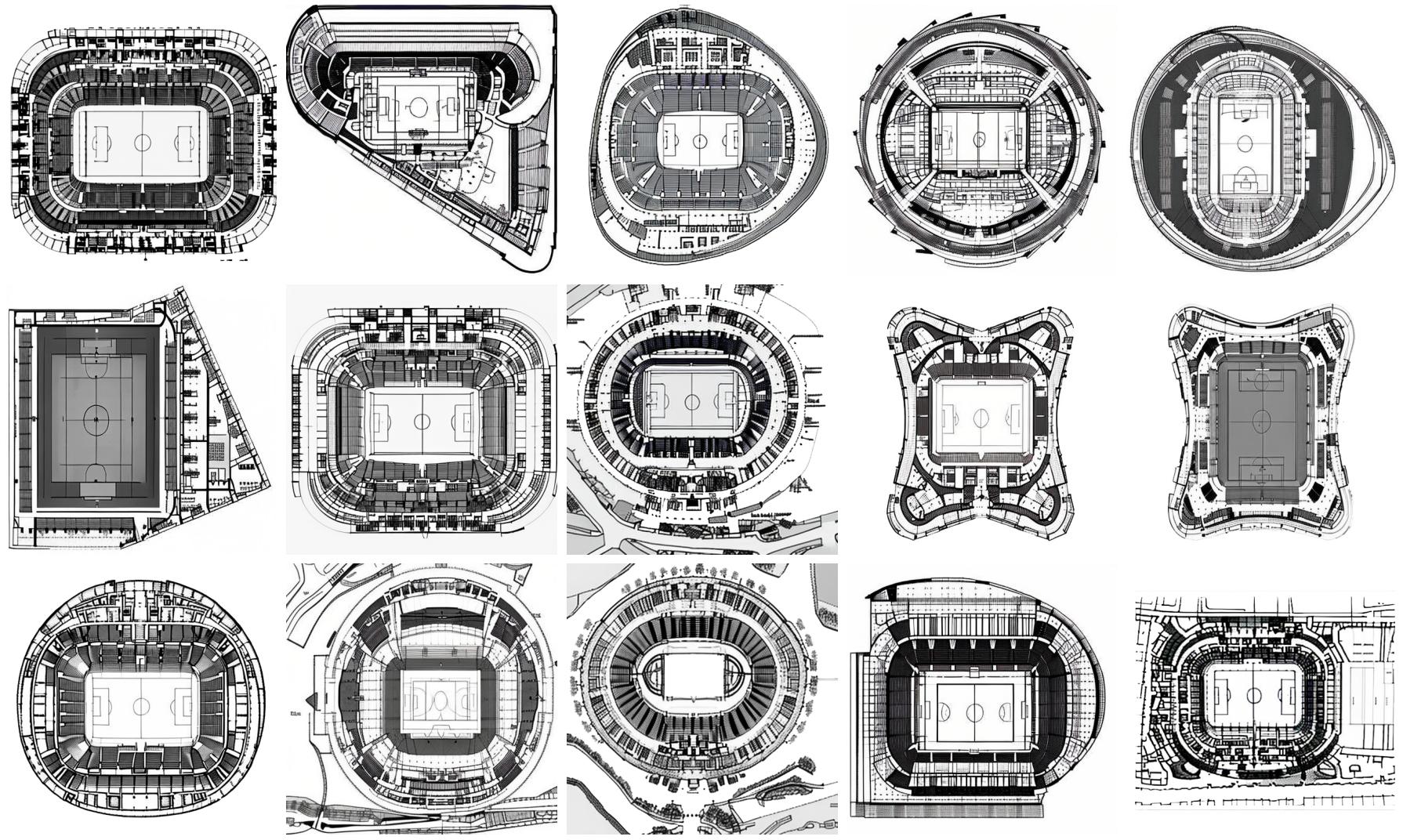}

   \caption{Generated floorplans for football stadium from various input forms and text prompt: 'a floor plan for a football stadium'.}
   \label{fig:stadium}
\end{figure}

\begin{figure}[ht]
  \centering

   \includegraphics[width=1\linewidth]{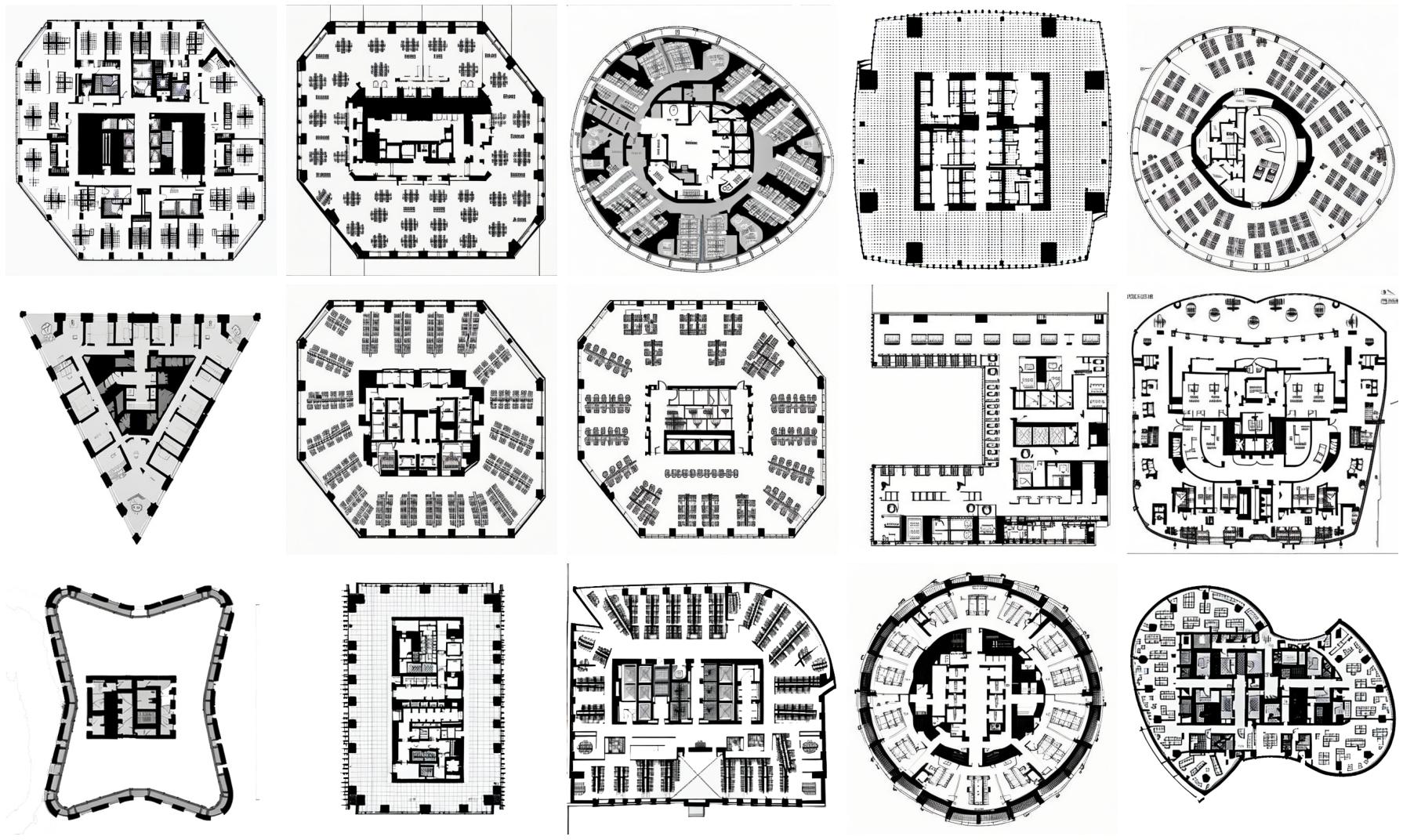}

   \caption{Generated floorplans for office buildings with one core.}
   \label{fig:office}
\end{figure}

\begin{figure}[ht]
  \centering

   \includegraphics[width=1\linewidth]{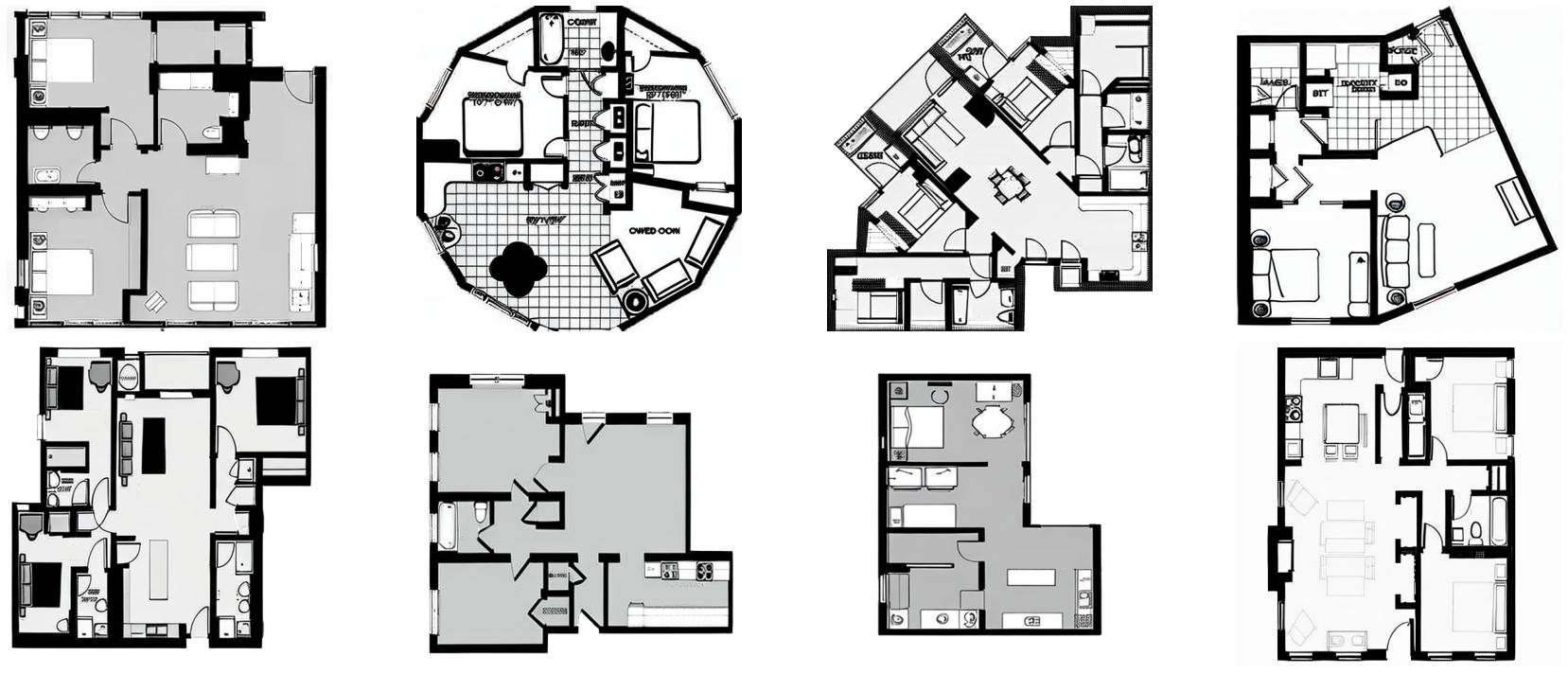}

   \caption{Generated floorplans for apartments.}
   \label{fig:apartment}
\end{figure}

\begin{figure}[ht]
  \centering

   \includegraphics[width=1\linewidth]{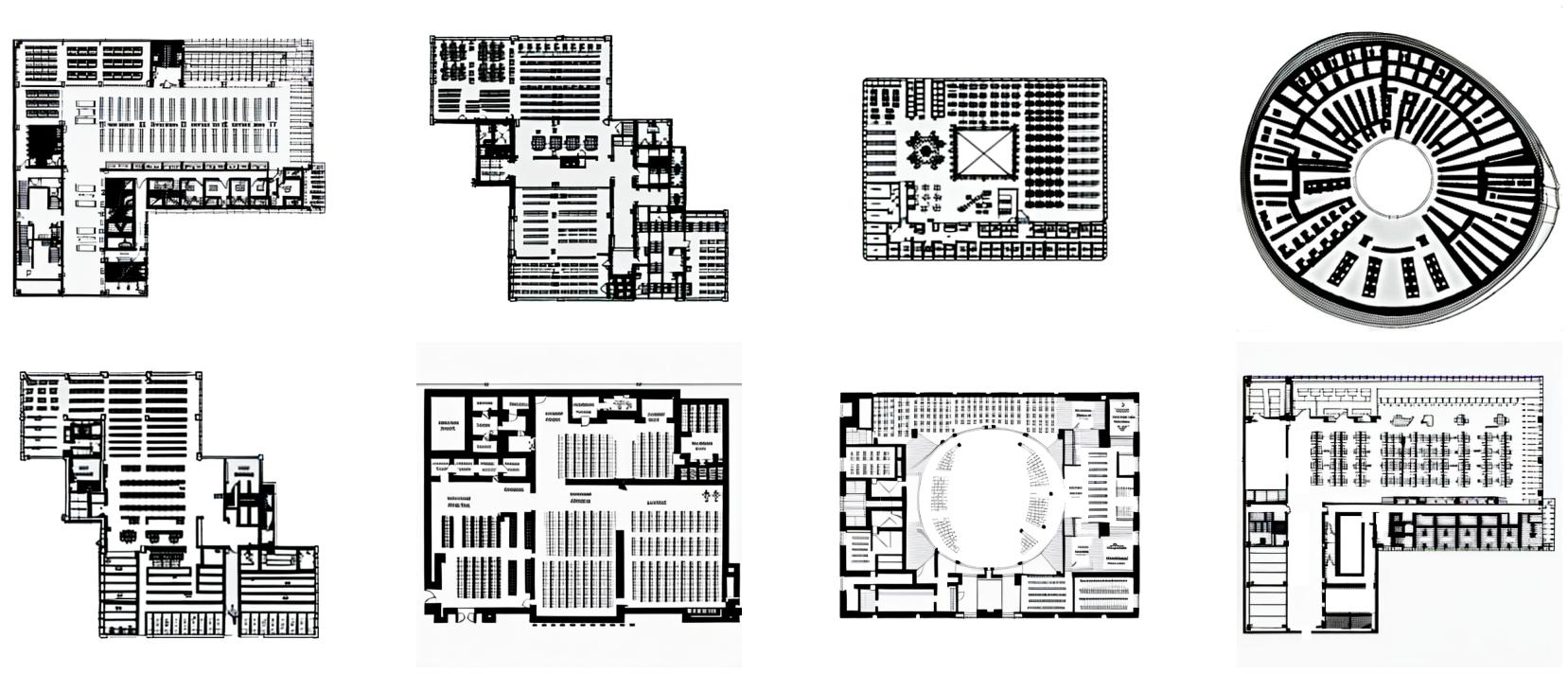}

   \caption{Generated floorplans for library buildings with various form inputs and text prompt: 'a floorplan for a library'.}
   \label{fig:library}
\end{figure}

\begin{figure}[ht]
  \centering

   \includegraphics[width=1\linewidth]{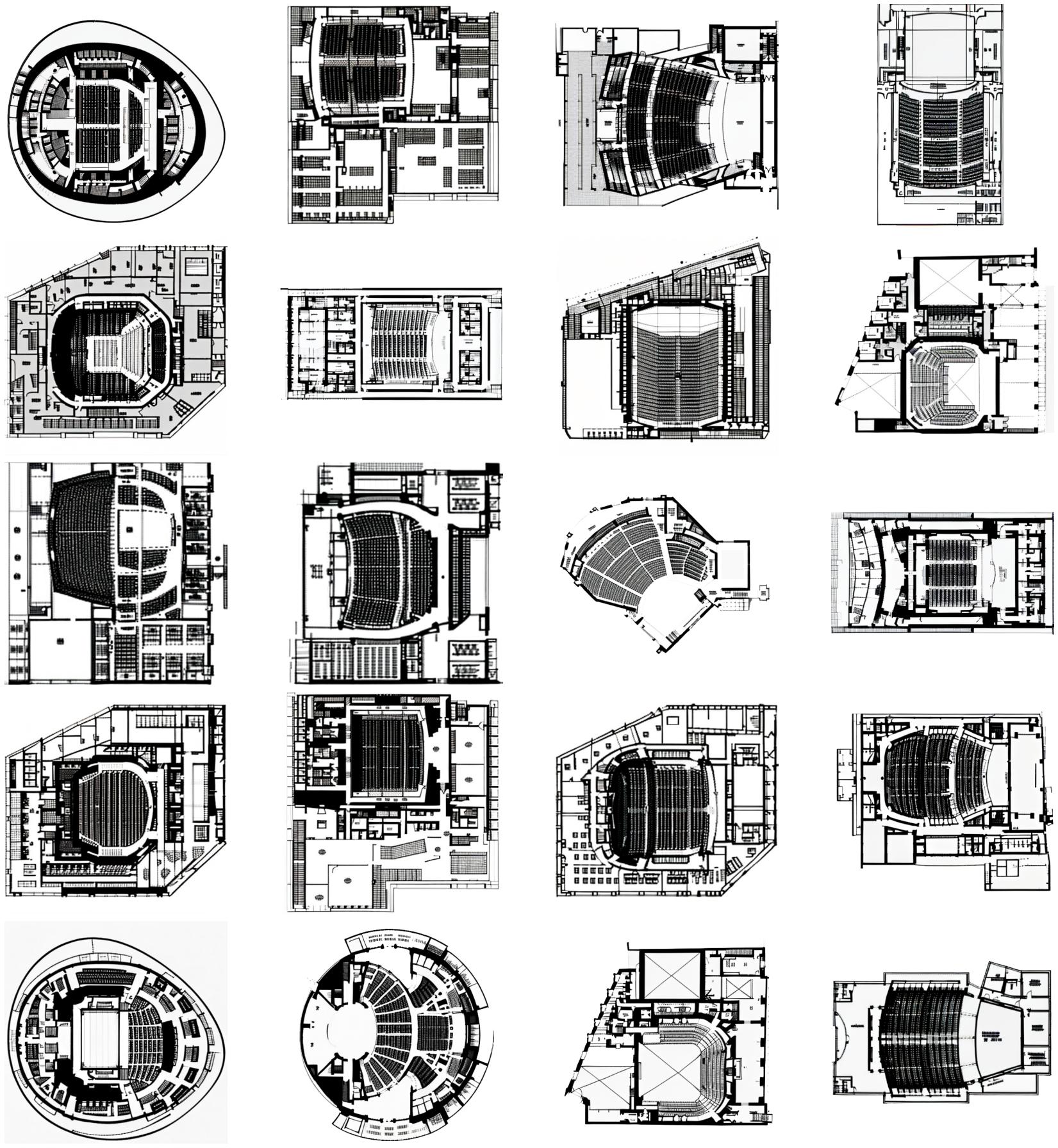}

   \caption{Generated floorplans for auditorium buildings with various form inputs and text prompt: 'a floorplan for an auditorium'.}
   \label{fig:auditorium}
\end{figure}

\textbf{Scale-agnostic approach:}
Undertaking a scale-agnostic approach, our introduced method transcends conventional limitations imposed by fixed scales. This innovative paradigm empowers our approach to seamlessly adapt to diverse scales, accommodating the design needs spanning from small-scale to large-scale building types (See Fig. \ref{fig:fig3}). The introduced method relies on a ratio-based driven approach, harnessing the inherent relationships between architectural components to guide the design process. Just as humans, when they learn to design, are able to adapt to different design briefs and building functionalities, our approach capitalises on the adaptability innate to the human design process. 

By leveraging the intrinsic proportions and ratios that underpin well-balanced designs across various buildings functionalities, we facilitate the creation of harmonious and aesthetically pleasing outcomes. In combination, these dynamic strategies not only enhance our model's flexibility but also imbue it with a heightened capacity to navigate the complexity of architectural design across varied scales and proportions. As a result, our methodology outperforms currently existing Generative floor plan tools which are predominantly confined to learn the complexity of a singular building functionality, such as housing floor plans. This broader purview expands our model's capabilities to encompass a more comprehensive range of architectural designs and functionalities.

\begin{figure}[ht]
  \centering

   \includegraphics[width=1\linewidth]{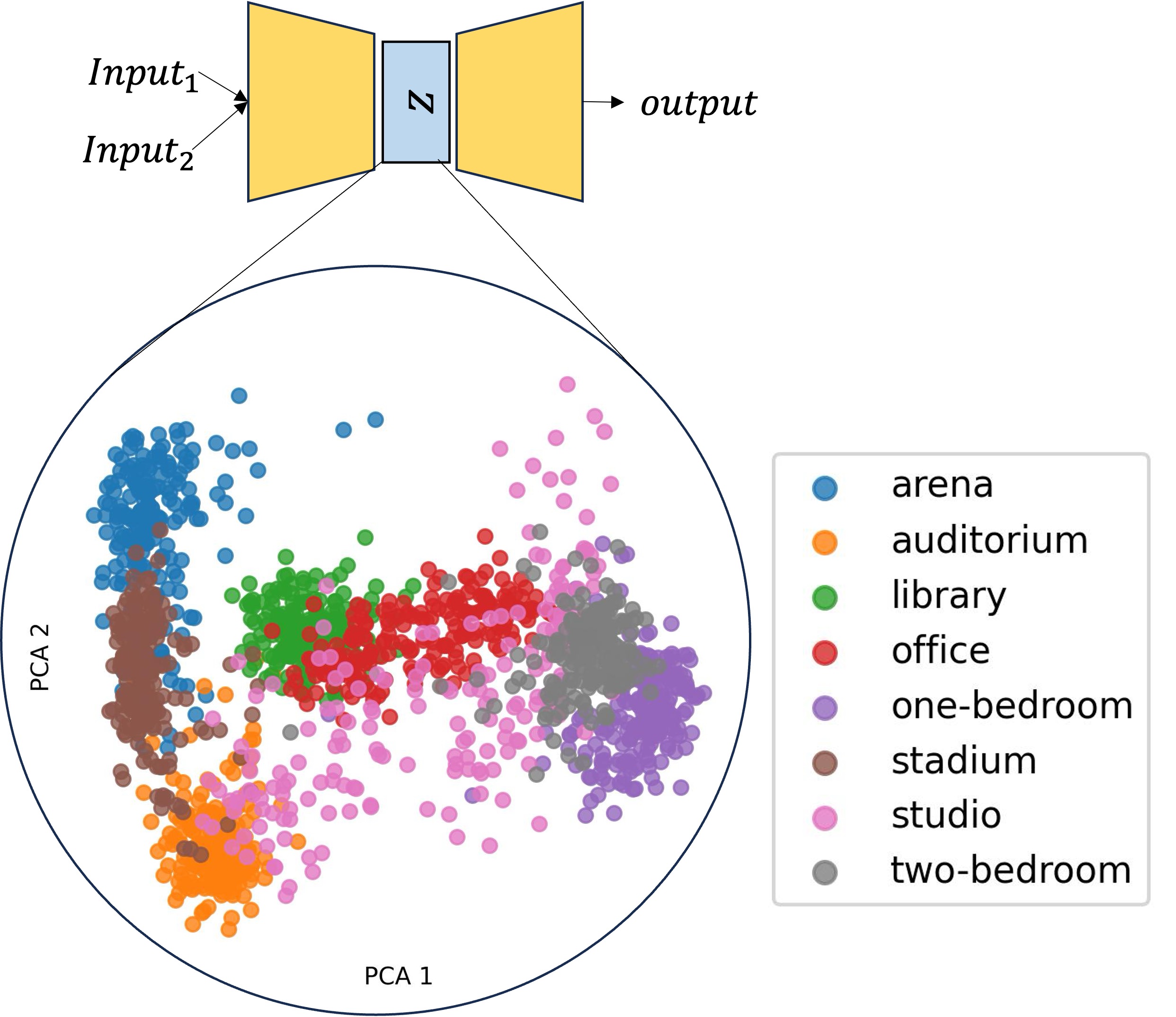}
   \caption{Visualising the introduced model's latent space (N=1600) showing the ability of the model to self-learn the proximity of different design components (i.e. the proximity of designing an arena to a stadium).}
   \label{fig:embeddings}
\end{figure}

\textbf{Latent space $(z_t)$ visualisation:}
To foster a deeper understanding of our model and advance the cause of explainable AI, we present a visualisation of our model's latent space $((z_t))$ in Fig. \ref{fig:embeddings}. This space is based on 1600 distinct generated outputs, resulting from various input forms and text prompts characterising different building functionalities. By applying Principal Component Analysis (PCA) to reduce the dimensionality of these embeddings—and without prior knowledge of specific input labels such as 'area', 'office', or 'stadium'—this figure demonstrates the model's capability to project embeddings that capture the nuances of building functionality. For example, it discerns the similarity between an arena's design and that of a stadium,  or between a one-bedroom and a two-bedroom apartment. It also observes the affinity between the designs of office buildings and libraries. Furthermore, the model presents the design of a studio (a singular room) as a foundational design, exhibiting proximity to various other building functionalities. Such insights emphasise the model's exceptional skill in data representation and its inherent sense of the relational closeness among diverse design principles.

\begin{figure}[ht]
  \centering

   \includegraphics[width=1\linewidth]{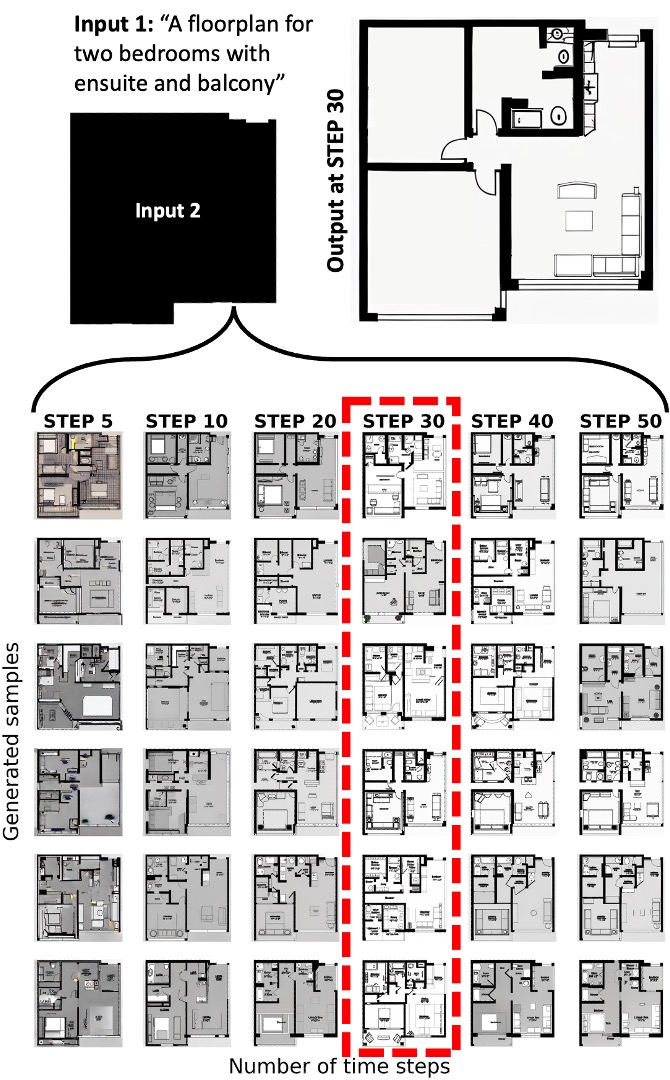}

   \caption{Image fidelity and denoising steps}
   \label{fig:fig4}
\end{figure}

\textbf{Image fidelity and denoising steps:} 
Figure \ref{fig:fig4} sheds light on the intricate connection between image fidelity and the denoising steps inherent in our model. These steps guide the transition from noisy starting points to polished images. The number of steps plays a pivotal role, impacting the depth of transformations. An increased number of steps permits the capture of finer details and intricate structures, elevating image fidelity. However, excessive steps can introduce challenges such as overfitting and a loss of diversity, potentially yielding hyper-realistic but less diverse images. Thus, striking an optimal balance is paramount, ensuring that the generated images possess authenticity, richness, and diversity reminiscent of real-world visual datasets.

\begin{figure}[ht]
  \centering

   \includegraphics[width=1\linewidth]{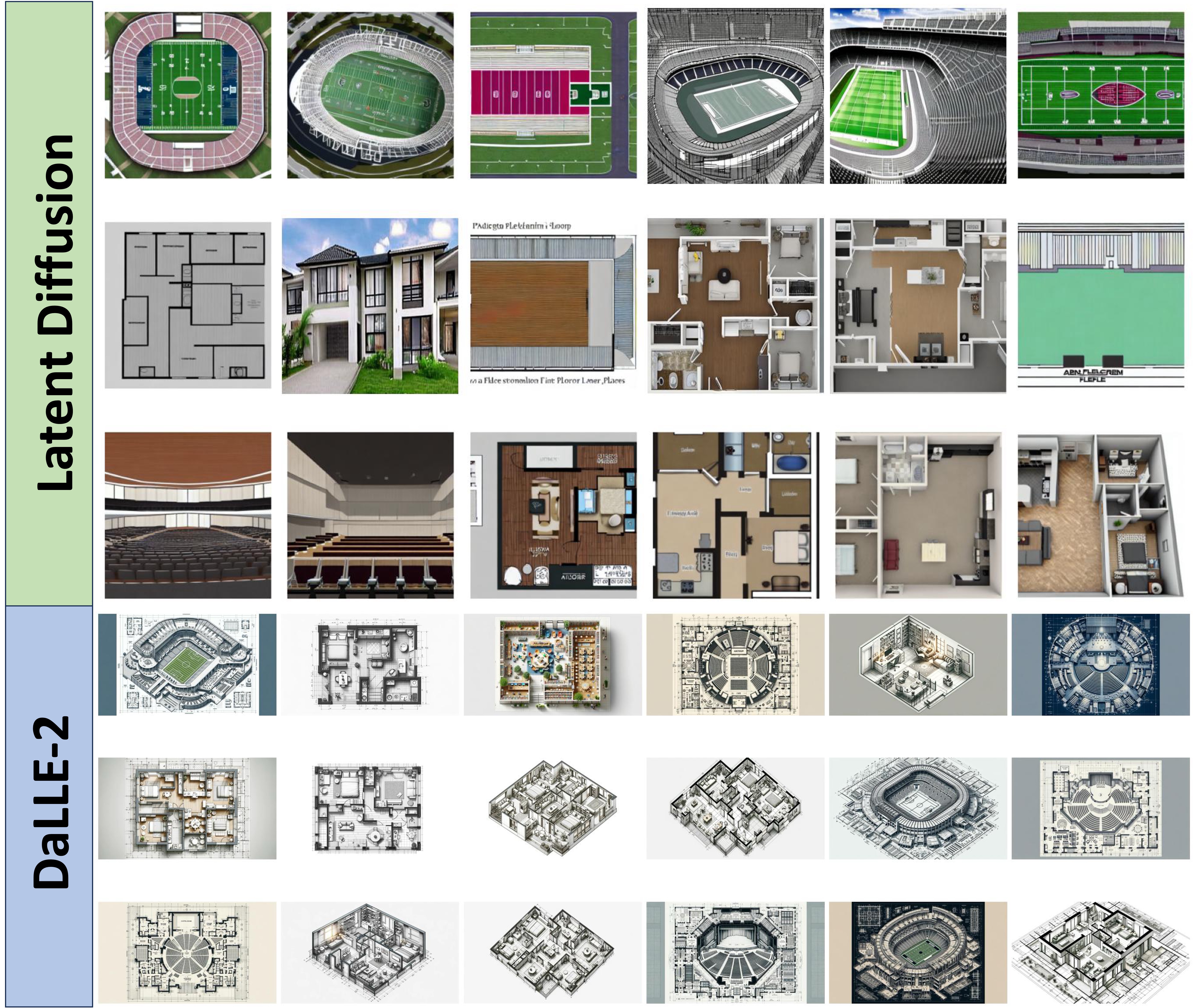}

   \caption{Generated images from the \textbf{state-of-the-art models: LDM \cite{latent_diffusion} and DaLLE-2}  using our prompts. For football stadium: 'a floorplan for a football stadium', for an apartment: 'a floorplan for a two bedroom apartment' and 'a floorplan for studio apartment', and for auditorium: 'a floorplan for an auditorium'.}
   \label{fig:stablediffusion_results}
\end{figure}

\textbf{Comparing the results to baselines:}
In Fig. \ref{fig:stablediffusion_results}, we present a  comparative analysis, juxtaposing the outcomes of our model against one of the state-of-the-art text-to-image foundational models and cutting-edge AI tools. This comparison highlights the existing challenges in generating floorplans, specifically regarding inconsistencies in quality and style.  The significance of our contributions is empirically underscored by this  demonstration. The ability to consistently imbue floorplans with a distinct architectural style not only sets our model apart but also accentuates its pioneering position within the landscape of state-of-the-art tools. The results establish the substantial stride our work takes in advancing the field, surpassing existing benchmarks and reshaping the trajectory of floorplan synthesis for different building’s types.

\section{Conclusion}
In this paper, we introduce a method for an AI-generative floorplan driven by a latent diffusion model. Unlike traditional architectural design processes, which are labour-intensive and rely heavily on architectural expertise, our model redefines floorplan creation. It not only replicates existing designs but also generates entirely new configurations by learning from diverse building types and architectural design variations. A key feature of our model is its inherent scale adaptability, effortlessly adjusting floorplan scale based on input footprints. This adaptability enhances versatility, enabling architects to design structures of varying sizes and footprint, from single houses to complex public buildings, without scale constraints. Importantly, our model surpasses the limitations of existing AI models for generated floorplans, which predominantly focus on residential functionalities. In contrast, our model extends its capabilities to various functionalities, providing architects with a comprehensive toolset for a wide range of design scenarios. Our research empowers architects to transcend traditional constraints, explore novel design solutions, and address diverse architectural needs.

{
    \small
    \bibliographystyle{ieeenat_fullname}
    \bibliography{main}
}


\clearpage
\onecolumn

%% Supplementary Section
\newpage
\section{Supplementary Material}

\subsection{Training and implementation details}
We followed closely the implementation of latent diffusion described by \cite{latent_diffusion, controlnet}. At the first stage of our method, we leveraged the existing pretrained weights for a diffusion model trained on LAION dataset \cite{laion400m} for text-to-image generation \cite{latent_diffusion}.  At the second stage, there are several training procedures that we undertook to synergise image, building form input, and textual prompt as follows:

\textbf{Generator implementations: } We used a Spatial Control Mechanism to infuse spatial inductive biases into our framework. With spatial transformers, it modulates image features based on: $x' = Ax + t$ where \( x' \) is the adjusted position, \( x \) is the original, \( A \) is the transformation matrix, and \( t \) is the translation vector. We combined the U-Nets with additional  transformer layers as previously described in methodology and in our experiments we used one layer to enhance the model's contextual understanding based on the described attention mechanism.  During training the images are processed at a resolution of 64x64 and upsampled to 512X512 at inference. The architecture further ensures that the conditioning stage remains non-trainable, utilising cross-attention based on the conditioning building's form. We monitoried performance via  validation loss metric. Diving deeper into the control mechanism, the ControlNet is configured to accept inputs with 4 channels, complemented by a 3 channel text prompts. This network harnesses 320 model channels, with attention mechanism designed across resolutions [4, 2, 1] and a transformer depth of 1. 

\textbf{Latent Space Modelling: } To model the generative latent space, we used Autoencoder architecture introduced by \cite{controlnet} to project multi-modal data into a unified latent space, characterised by encoding function \( f(x) \) and decoding mechanism \( g \), its primary objective defined as: $L(x, g(f(x)))$ ensuring the latent representation's fidelity to the original data. The Autoencoder undertakes initial processing, embracing an embedding dimension of 4 and operating at 256x256 resolution with a layered approach of 3 input channels, 4 latent space z-channels, and 3 output channels for the generated image.

\textbf{Textual Embeddings:} Leveraging the robustness of the OpenCLIP model \cite{openclip}, we createdd a frozen OpenCLIP embedder to transfrom texts into textual embeddings. This pre-trained module assists in integrating textual content into the introduced framework, ensuring the embedding mechanism remains static during the training process.

\textbf{Training Paradigm: } Trained models are resumed from pre-trained checkpoints of the first-stage. The training regimen is executed with 32-bit precision, with batch size 1, and training cycles 429 epochs. We used Adam \cite{adam} optimiser to train our model with a  learning rate that progresses linearly across timesteps. Specifically, the learning rate is initiated at a rate of 0.00085 and culminating at 0.0120. Over its span of 1000 timesteps, periodic logs are captured every 200 timesteps.

\subsection{Floorplan evaluation tool}

Fig. \ref{fig:game} shows the interface of the developed tools for evaluating generated and real floorplan designs based on experts knowledge. To evaluate floorplans, the link to the game: \href{https://floorplan.streamlit.app}{FloorplanGame} 

\begin{figure*}[!t]
  \centering

   \includegraphics[width=1\linewidth]{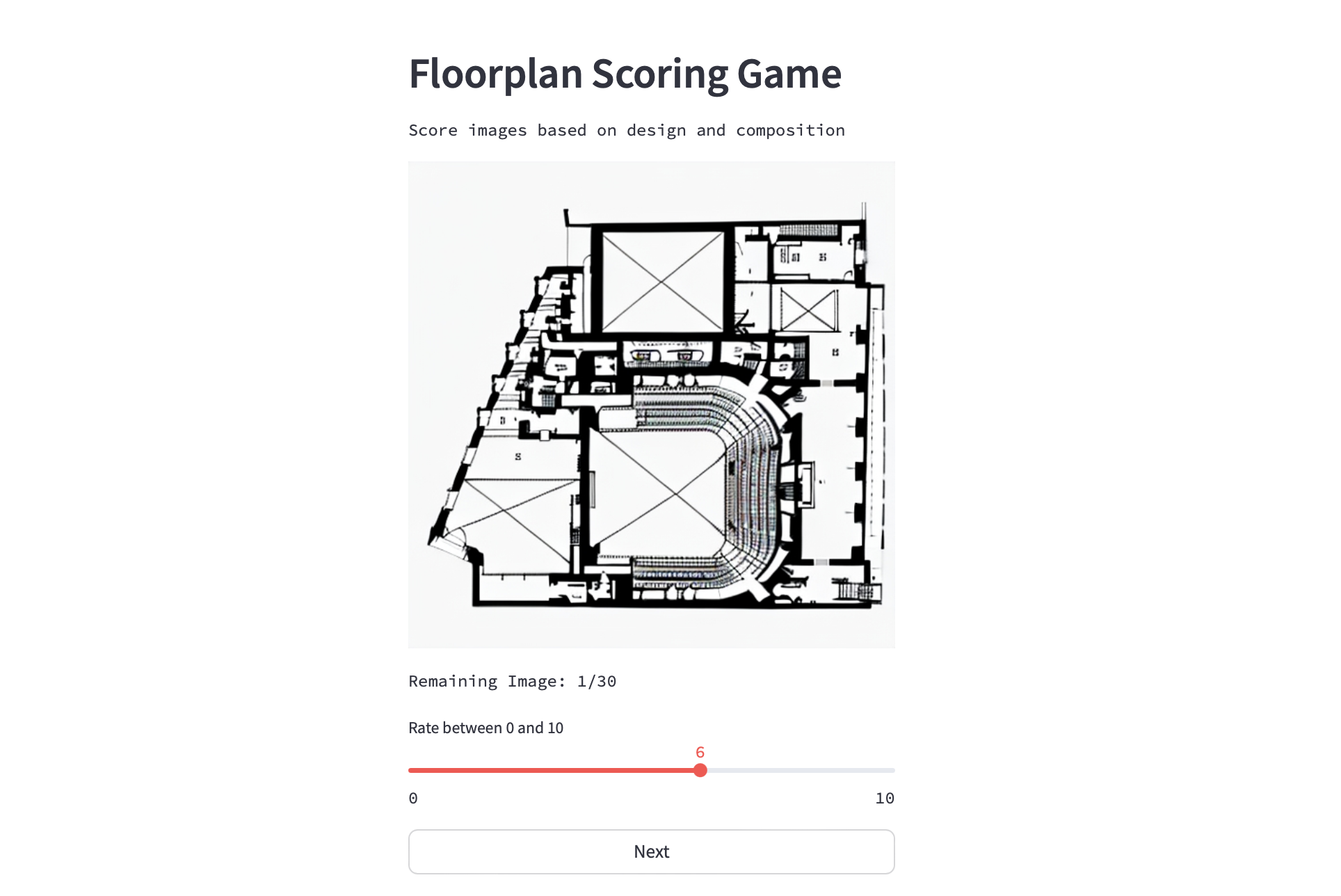}

   \caption{The interface for the developed Floorplan Evaluation tool.}
   \label{fig:game}
\end{figure*}

\subsection{High-resolution images}

 Fig. \ref{fig:stadium},  \ref{fig:office}, \ref{fig:apartment}, \ref{fig:library}, and \ref{fig:auditorium} show the generated designs (in a higher resolution) for various buildings footprints for stadium, office, apartment, library and auditorium buildings respectively highlighting the complexity and diversity of the generated design alternatives.

\begin{figure*}[h]
  \centering

   \includegraphics[width=1\linewidth]{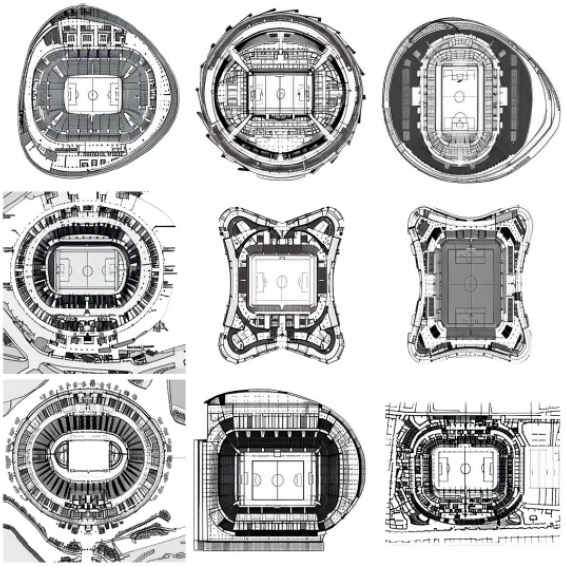}

   \caption{Generated floorplans for football stadium from various input forms and text prompt: 'a floor plan for a football stadium'.}
   \label{fig:stadium}
\end{figure*}

\begin{figure*}[h]
  \centering

   \includegraphics[width=1\linewidth]{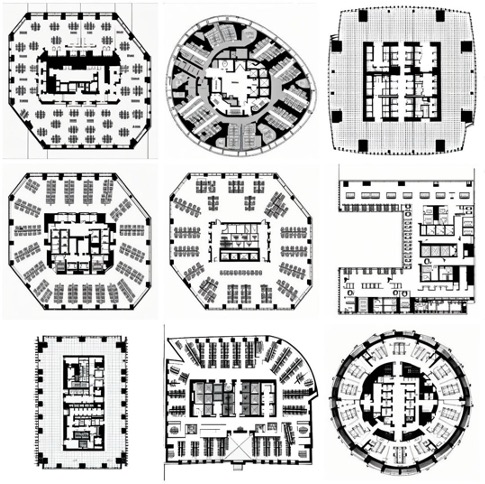}

   \caption{Generated floorplans for office buildings with one core.}
   \label{fig:office}
\end{figure*}

\begin{figure*}[h]
  \centering

   \includegraphics[width=1\linewidth]{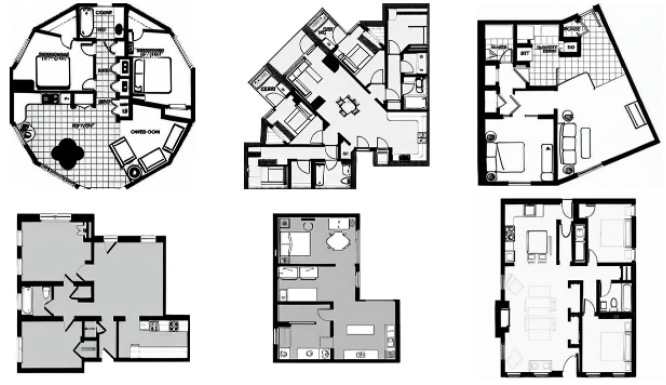}

   \caption{Generated floorplans for apartments.}
   \label{fig:apartment}
\end{figure*}

\begin{figure*}[h]
  \centering

   \includegraphics[width=1\linewidth]{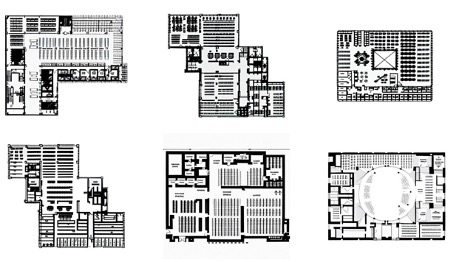}

   \caption{Generated floorplans for library buildings with various form inputs and text prompt: 'a floorplan for a library'.}
   \label{fig:library}
\end{figure*}

\begin{figure*}[h]
  \centering

   \includegraphics[width=1\linewidth]{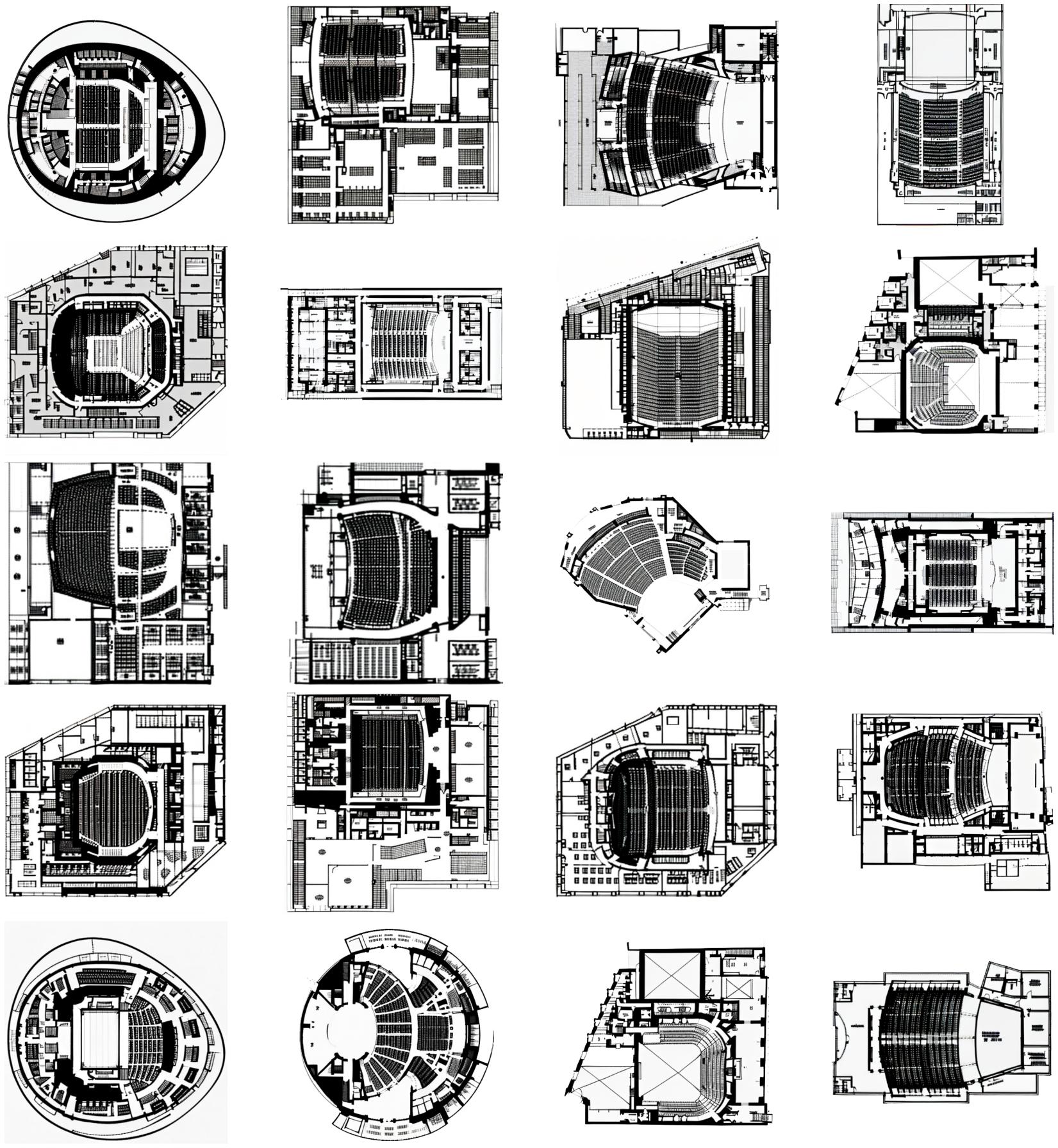}

   \caption{Generated floorplans for auditorium buildings with various form inputs and text prompt: 'a floorplan for an auditorium'.}
   \label{fig:auditorium}
\end{figure*}

\subsection{Prompt engineering}

Prompt engineering plays a pivotal role in elevating the quality of generated images through precise and strategic manipulation of input prompts \cite{prompt_1, prompt_2, prompt_3}. This technique involves fine-tuning the textual or input prompts provided to generative models, with the aim of steering the output towards desired visual outcomes. By carefully crafting prompts that encapsulate specific visual details, styles, or attributes, users can effectively guide the model's creative process.
By tailoring prompts to emphasize certain features, textures, or compositions, the generative model becomes more attuned to the creator's intent. Consequently, the resulting images exhibit heightened fidelity to the envisioned aesthetics. Based on your experiments, we found that even subtle modifications to the provided prompt can wield over the quality of the generated image. our experiments showed that simple adjustments to the wording, context, or emphasis within the prompt can trigger nuanced shifts in the generative model's output. By crafting prompts that encapsulate not only visual specifics but also stylistic preferences, the user can effectively guide the model's interpretive process. As a result, the generated images exhibit an elevated level of alignment with the desired aesthetics.

\end{document}